\documentclass{llncs}
\usepackage{comment}
\usepackage{url}
\usepackage{amssymb} 
\usepackage{amsmath} 
\usepackage{paralist}
\usepackage{stmaryrd} 
\usepackage{url}
\usepackage{multirow} 
\usepackage{times}
\usepackage{helvet}
\usepackage{courier}
\usepackage{graphicx}
\usepackage[linesnumbered]{algorithm2e}
\usepackage{algorithmic}
\usepackage{hyperref}
\usepackage{caption}

\begin{document}

\title{A Distance-based Paraconsistent Semantics for DL-Lite}
\author{Xiaowang Zhang$^1$ \and Kewen Wang$^2$ \and Zhe Wang$^2$ \and Yue Ma$^3$ \and Guilin Qi$^4$}
\institute{$^1$School of Computer Science and Technology, Tianjin University, China\\
$^2$School of Information and Communication Technology, Griffith University, Australia\\
$^3$Laboratoire de Recherche en Informatique, University Paris Sud, France\\
$^4$School of Computer Science and Engineering, Southeast University, China}

\maketitle
\begin{abstract}
DL-Lite is an important family of description logics. Recently, there is an increasing interest in handling inconsistency in DL-Lite as the constraint imposed by a TBox can be easily violated by assertions in ABox in DL-Lite. In this paper, we present a distance-based paraconsistent semantics based on the notion of feature in DL-Lite, which provides a novel way to rationally draw meaningful conclusions even from an inconsistent knowledge base. Finally, we investigate several important logical properties of this entailment relation based on the new semantics and show its promising advantages in non-monotonic reasoning for DL-Lite.
\end{abstract}

%
%

\section{Introduction}\label{sec:intro}

The DL-Lite \cite{Artale2009} is a family of lightweight description logics (DLs), the logical foundation of OWL 2.0 QL, one of the three profiles of OWL 2.0 for Web ontology language recommended by W3C \cite{DBLP:journals/ws/GrauHMPPS08}. In description logics, an ontology is expressed as a knowledge base (KB).
Inconsistency is not rare in ontology applications and can be caused by several reasons, such as errors in modeling, migration from other formalisms, ontology merging, and ontology evolution. In the age of big data, it is becoming impossible to avoid inconsistency of larger and larger scale of KBs. Therefore, handling inconsistency
is always considered an important problem in DLs and ontology
management communities \cite{DBLP:conf/semweb/RosatiRGM11}. However, DL-Lite reasoning mechanism based
on classical DL semantics faces problem when inconsistency occurs, which is referred to as the triviality problem. That is, any
conclusions, that are possibly irrelevant or even contradicting,
will be entailed from an inconsistent DL-Lite ontology
under the classical semantics.

In many practical ontology applications, there is a strong need for inferring (only) useful information from inconsistent ontologies. For instance, consider a simple DL-Lite KB $\mathcal{K}=(\mathcal{T},\mathcal{A})$  where $\mathcal{T}=\{\textit{Penguin}
\sqsubseteq \textit{Bird}$, $\textit{Swallow} \sqsubseteq \textit{Bird}$, $\textit{Bird} \sqsubseteq \textit{Fly}\}$ and $\mathcal{A}=\{\textit{Penguin}(\textit{tweety})$, $\neg \textit{Fly}(\textit{tweety})$, $\textit{Swallow}(\textit{fred})\}$. That KB tells us that penguins are birds; swallows are birds; birds can fly; \textit{tweety} is a penguin; \textit{tweety} cannot fly; and \textit{fred} is a swallow. Under the classical semantics for DLs, anything can be inferred from $\mathcal{K}$ since $\mathcal{K}$ is not inconsistent (i.e., it has no any model.). Intuitively, one might wish to
still infer $\textit{Bird}(\textit{fred})$ and $\textit{Fly}(\textit{fred})$, while it is useless to derive both $\textit{Fly}(\textit{tweety})$ and $\neg \textit{Fly}(\textit{tweety})$ from $\mathcal{K}$.

There exist several proposals for reasoning
with inconsistent DL-Lite KBs in the literature. These approaches usually fall into one of two fundamentally different streams. The first one is based on the assumption that inconsistencies are caused by erroneous data and thus, they should be removed in order to obtain a consistent KB
\cite{DBLP:conf/esws/KalyanpurPSG06,DBLP:conf/aaai/MeyerLBP06,DBLP:conf/semweb/DolbyFFKKMMSW07,DBLP:conf/www/DuS08}. In most approaches in this stream, the task of repairing inconsistent ontologies is actually reduced to finding a maximum consistent subset of the original KB. A shortcoming of these approaches is similar to the so-called \emph{multi-extension problem} in Reiter's default logic. That is, in many cases, an inconsistent KB may have several different sub-KBs
that are maximum consistent. The other stream, based on the idea of living with inconsistency, is to introduce a form of paraconsistent reasoning or inconsistency-tolerant reasoning by employing non-standard reasoning methods (e.g., non-standard inference and non-classical semantics). There are some strategies to select consistent subsets from an inconsistent KB as substitutes of the original KB in reasoning \cite{DBLP:conf/ijcai/SchlobachC03,DBLP:conf/ijcai/HuangHT05,DBLP:conf/rr/LemboLRRS10,DBLP:journals/logcom/Kamide12,DBLP:journals/igpl/HuangLH13,DBLP:journals/jiis/ZhangL13}. The Belnap's four-valued semantics has been successfully extended into DL-Lite \cite{DBLP:conf/rr/MaH09} where two additional logical values besides ``true" and ``false" are introduced to indicate contradictory conclusions. Inference power of the four-valued semantics is further enhanced by a new quasi-classical semantics for DLs proposed by Zhang et~al. \cite{DBLP:journals/ijar/ZhangXLJ14}, which is a generalization of Hunter's quasi-classical semantics for propositional logic. However, the reasoning capability of such paraconsistent methods is not strong enough for many practical applications. For instance, a conclusion $\phi$, that can inferred from a consistent KB $\mathcal{K}$ under the classical semantics, may become not derivable under their paraconsistent semantics. We argue that approaches in these two streams are mostly \emph{coarse-grained} in the sense that they fail to fully utilize semantic information in the given inconsistent KB. For instance, when two interpretations make a concept unsatisfiable, one interpretation may be more reasonable than the other. But existing approaches to paraconsistent semantics in DLs do not take this into account usually.

Recently a distance-based semantics presented by Arieli \cite{DBLP:journals/ijar/Arieli08} has been proposed to deal with inconsistent KBs in propositional logic. However, it is not straightforward to generalize this approach to DLs because it directly works on models (it is feasible in propositional logic since a propositional KB has a finite number of finite models) while, in DLs, a KB might have infinite number of models and a model might also be infinite. Additionally, it is also a challenge in adopting distance-based semantics for complex constructors in DLs.

To overcome these difficulties, in this paper we first use the notion of
\emph{features} \cite{DBLP:conf/aaai/WangWT10} and introduce a distance-based semantics for paraconsistent reasoning with DL-Lite. Features in DL-Lite are Herbrand interpretations extended with limited structure, which provide a novel semantic characterization for DLs. In addition, features also generalize the notion of \emph{types} for TBoxes \cite{DBLP:conf/kr/KontchakovWZ08} to general KBs.
Each KB in DL-Lite has a finite number of features and each feature is finite. This makes it possible to cast Arieli's distance-based semantics to DL-Lite.

The main innovations and contributions of this paper can be summarized as follows. We introduce distance functions on \emph{types} of DL-Lite$^{\mathcal{N}}_{bool}$ KBs, which avoids the problem of domain infiniteness and model infiniteness in defining the distance function in terms of models of KBs. We choose DL-Lite$^{\mathcal{N}}_{bool}$ \cite{Artale2009}, one of the most expressive members of the DL-Lite family, and define distance-based semantics for DL-Lite$^{\mathcal{N}}_{bool}$ in a way analogous to the model-based approaches in propositional logic. Although our approach is based on DL-Lite$^{\mathcal{N}}_{bool}$, we argue that our technique can easily be adapted to other DLs.
Based on the new distance function on types, we develop a way of measuring types that are closest to a TBox and the notion of \emph{minimal model types} is introduced. This notion is also extended to \emph{minimal model features} for KBs. We propose a distance-based semantics for DL-Lite$^{\mathcal{N}}_{bool}$ so that useful information can still be inferred when a KB is inconsistent. This is accomplished by introducing a novel entailment relation (i.e. distance-based entailment) between a KB and an axiom in terms of minimal model features. Our results show that the distance-based entailment is paraconsistent, non-monotonic, cautious as the paraconsistent based on multi-valued semantics. We also show that the distance-based entailment is not over-skeptical in the sense that for a classically consistent KB, the distance-based entailment coincides with the classical entailment, which is missing in most existing paraconsistent semantics for DLs.
%
Due to the space limitation, all proofs are omitted but they are available in an extended technical report in \cite{DBLP:journals/corr/abs-1301-2005}.

\section{The DL-Lite family and features}\label{sec:pre}
\paragraph{\textbf DL-Lite $^{\mathcal{N}}_{bool}$} \quad
A \textit{signature} is a finite set $\Sigma = \Sigma_A \cup \Sigma_R
\cup \Sigma_I \cup \Sigma_N$ where $\Sigma_A$ is the set of atomic
concepts, $\Sigma_R$ the set of atomic roles, $\Sigma_I$ the
set of individual names (or, objects) and $\Sigma_N$ the set of
natural numbers in $\Sigma$. We use capital letters $A,B,C$ ( with subscripts $C_{1},C_{2}$) to denote concept names, $P, R, S$ (with subscripts $P_{1},P_{2}$) to denote role names, lowercase letters $a,b,c$ to denote individual names and assume 1 is always in $\Sigma_N$. $\top$ and $\bot$ will not be considered as concept names or role names. 

Formally, given a signature $\Sigma$, the DL-Lite$^{\mathcal{N}}_{bool}$ language is inductively constructed by syntax rules: (1) $R \leftarrow P\mid P^-$; (2) $B\leftarrow\top \mid A\mid \geq n R$;  and (3) $C\leftarrow B \mid \neg C\mid C_1\sqcap C_2$. We say $B$ a \textit{basic concept} and $C$ a \textit{general concept}. Other standard concept constructs such as $\bot$, $\exists R$, $\leq n R$ and $C_{1}\cup C_{2}$ can be introduced as abbreviations: $\bot$ for $\neg \top$, $\exists R$ for $\geq 1 R$, $\leq n R$ for $\neg (\geq (n+1) R)$ and  $C_{1}\sqcup C_{2}$ for $\neg (\neg C_{1}\sqcap \neg C_{2})$. For any $P\in \Sigma_{R}$, $P^{--}=P$.

A TBox $\mathcal{T}$ is a finite set of \textit{(concept) inclusions}  of the form $C_1\sqsubseteq C_2$ where $C_1$ and $C_2$ are general concepts. An ABox $\mathcal{A}$ is a finite set of concept assertions $C(a)$ and role assertions $R(a,b)$. Concept inclusions, concept assertions and role assertions are axioms. A KB is composed of a TBox and an ABox, written by $\mathcal{K}=(\mathcal{T},\mathcal{A})$. $Sig(\mathcal{K})$ denotes the signature of $\mathcal{K}$.

An interpretation $\mathcal{I}$ is a pair $\langle
\Delta^\mathcal{I},\cdot^\mathcal{I}\rangle$, where
$\Delta^\mathcal{I}$ is a non-empty set called the \textit{domain} and
$\cdot^\mathcal{I}$ is an interpretation function such that $a^\mathcal{I}\in\Delta^\mathcal{I}$, $A^\mathcal{I}\subseteq\Delta^\mathcal{I}$ and $P^\mathcal{I}\subseteq \Delta^\mathcal{I}\times\Delta^\mathcal{I}$. General concepts are interpreted as follows: $ (P^-)^\mathcal{I} = \{(a^\mathcal{I}, b^\mathcal{I})\mid\, (b^\mathcal{I},a^\mathcal{I})\in P^\mathcal{I}\}$,  $(\geq n R)^\mathcal{I} = \{a^\mathcal{I}\mid\, |\{b^\mathcal{I}\mid\, (a^\mathcal{I},b^\mathcal{I})\in R^\mathcal{I}\}| \geq n\} $, $(\neg C)^\mathcal{I} = \Delta^\mathcal{I}\setminus C^\mathcal{I} $, and $ (C_{1}\sqcap C_{2})^\mathcal{I} = C_{1}^\mathcal{I}\cap C_2^\mathcal{I}$. The definition of interpretation is based on the \textit{unique name assumption} (UNA), i.e., $a^\mathcal{I}\neq b^\mathcal{I}$ for two different individual names $a$ and $b$.

An interpretation $\mathcal{I}$ is a \textit{model} of
a concept inclusion $C_1\sqsubseteq C_2$ (a concept assertion $C(a)$, or a role assertion $R(a,b)$) if $C_1^\mathcal{I}\subseteq C_2^\mathcal{I}$ ($a^\mathcal{I}\in C^\mathcal{I}$, or  $(a^\mathcal{I},b^\mathcal{I})\in R^\mathcal{I}$); and $\mathcal{I}$ is called a \textit{model} of a
TBox $\mathcal{T}$ (an ABox $\mathcal{A}$) if $\mathcal{I}$ is a model of each inclusion of $\mathcal{T}$ (each assertion of $\mathcal{A}$). $\mathcal{I}$ is called a \textit{model} of a KB
$(\mathcal{T},\mathcal{A})$ if $\mathcal{I}$ is a model of both
$\mathcal{T}$ and $\mathcal{A}$. We use $Mod(\mathcal{K})$ to denote the set
of models of $\mathcal{K}$. 
A KB $\mathcal{K}$ \textit{entails} an axiom $\phi$, if $Mod(\mathcal{K})\subseteq Mod(\{\phi\})$. Two KBs $\mathcal{K}_1$ and $\mathcal{K}_2$ are \textit{equivalent} if
$Mod(\mathcal{K}_1)=Mod(\mathcal{K}_2)$, denoted by
$\mathcal{K}_1\equiv \mathcal{K}_2$. A  KB $\mathcal{K}$
is \textit{consistent} if it has at least one mode, \textit{inconsistent} otherwise.

%

\paragraph{\textbf Features} \quad
Let $\Sigma$ be a signature. A $\Sigma$-\textit{type} (or simply a \textit{type}) is a set of basic concepts over $\Sigma$, s.t., $\top\in\tau$, and for any $m,n\in \Sigma_N$ with $m<n$, $R\in \Sigma_{R}\cup \{P^{-}\mid P\in \Sigma_{R}\}$, $\geq n R \in \tau$ implies $\geq m R\in \tau$. As $\top\in \tau$ for any type $\tau$, we omit it in examples for simplicity. $T_\Sigma$ denotes the set of all $\Sigma$-types. Note that if $\exists P$ (or $\exists P^-$) occurs in a general concept $C$ then $\exists P^-$ (or $\exists P$) should be also considered as a new concept independent of $\exists P$ (or $\exists P^-$) in computing types of $C$ respectively. We say a \textit{type set} as a set of types $\{\tau_{1},\ldots,\tau_{m}\}$, denoted as $\Xi$ and a \textit{type group} as a set of type sets $\{\Xi_1,\ldots,\Xi_n\}$, denoted as $\Pi$. Then we denote $\cup \Xi=\tau_{1}\cup\dots\cup \tau_{m}$ and $\cap\Pi=\Xi_1\cap\ldots\cap\Xi_n$.
%
%
%
%
A type $\tau$ \textit{satisfies} a basic concept $B$ if $B\in \tau$, $\tau$ satisfies $\neg C$ if $\tau$ does not satisfy $C$, and $\tau$ satisfies $C_{1}\sqcap C_{2}$ if $\tau$ satisfies both $C_{1}$ and $C_{2}$. $T_\Sigma(C)$ denotes a collection of all $\Sigma$-types of $C$. In this way, each general concept $C$ over $\Sigma$ corresponds to a set $T_{\Sigma}(C)$ of all $\Sigma$-types satisfying $C$. $\tau$ \textit{satisfies} a concept inclusion
$C\sqsubseteq D$ if $\tau\in T_\Sigma(\neg C\sqcup D)$. And $\tau$
is a \textit{model type} a TBox $\mathcal{T}$ iff it satisfies each
inclusion in $\mathcal{T}$. \textit{Model type sets} and \textit{model type groups} are analogously defined. If $\Xi$ is a model type set of a TBox $\mathcal{T}$ then $\exists P\in \cup \Xi$ iff $\exists P^{-} \in \cup \Xi$. This property is called \textit{role coherence} which can be used to check whether a type set is the model type set of some TBox. $\Pi_{\Sigma}(\mathcal{T})$ denotes the model type group $\{T_{\Sigma}(\neg C_1\sqcup D_1),\ldots,T_{\Sigma}(\neg C_n\sqcup D_n)\}$ of $\mathcal{T}$ where $\mathcal{T}=\{C_{1}\sqsubseteq D_{1},\ldots, C_{n}\sqsubseteq D_{n}\}$ is a TBox over $\Sigma$.
It appears that $\cap \Pi_{\Sigma}(\mathcal{T})$ is the collection of model $\Sigma$-types of $\mathcal{T}$. 
%

A $\Sigma$-\textit{Herbrand set} (or simply \textit{Herbrand set})
$\mathcal{H}$ is a finite set of member assertions satisfying: (1) for each $a\in \Sigma_I$, if $B_1(a),\ldots,B_k(a)$, where
$\{B_1,\ldots,B_k\}\subseteq \Sigma_B$ are all the concept
assertions about $a$ in $\mathcal{H}$, then the set $\{B_1,\ldots,
  B_k\}$ is a $\Sigma$-type; (2) for each $P\in \Sigma_R$, if $P(a,b_i)(1\leq i\leq n)$ are all the role assertions about $a$ in $\mathcal{H}$, then for any $m\in \Sigma_N$ with $m\leq n$, $(\geq m P) (a)$ is in $\mathcal{H}$; (3) for each $P\in \Sigma_R$, if $P(b_i,a)(1\leq i\leq n)$ are all the role assertions in $\mathcal{H}$, then for any $m\in \Sigma_N$ with $m\leq n$, $(\geq m P^-) (a)$ is in $\mathcal{H}$.
%
%
%
We simply write $\tau(a)=\{B_1(a),\ldots,B_k(a)\}$ where
$\tau=\{B_1,\ldots,B_k\}$. Moreover, given a set of types
$\Xi=\{\tau_1,\ldots,\tau_m\}$, $\Xi(a)$ denotes
$\{\tau_1(a),\ldots,\tau_m(a)\}$ without confusion. In this case, we
say $\tau(a)$ is in $\mathcal{H}$ if
$\{B_1(a),\ldots,B_k(a)\}\subseteq \mathcal{H}$.
%
A Herbrand set $\mathcal{H}$ \textit{satisfies} a concept assertion
$C(a)$ (a role assertion
$P(a,b)$ or $P^-(b,a)$) if $\tau(a)$ is in $\mathcal{H}$ and $\tau\in T_\Sigma(C)$ ($P(a,b)\in \mathcal{H}$ or  $P^-(b,a)\in \mathcal{H}$). A Herbrand set $\mathcal{H}$ \textit{satisfies} an ABox $\mathcal{A}$ if $\mathcal{H}$ satisfies all assertions in $\mathcal{A}$.

A $\Sigma$-\textit{feature} (or simply a \textit{feature}) $\mathcal{F}$
is a pair $\langle\Xi,\mathcal{H}\rangle$, where $\Xi$ is a non-empty
set of $\Sigma$-types and $\mathcal{H}$ a $\Sigma$-Herbrand set, if $\mathcal{F}$ satisfies: (1) for each $P\in \Sigma_R$, $\exists P\in \bigcup \Xi$ iff $\exists P^-\in \bigcup \Xi$ (i.e., $\Xi$ holds role coherence); and (2) for each $a\in \Sigma_I$ and $\tau(a)$ in $\mathcal{H}$,
  s.t., $\tau$ is a $\Sigma$-type, $\tau\in\Xi$.
A feature $\mathcal{F}$ \textit{satisfies} an inclusion
$C_1\sqsubseteq C_2$ over $\Sigma$, if $\Xi\subseteq T_\Sigma(\neg
C_1\sqcup C_2)$; $\mathcal{F}$ \textit{satisfies} a concept assertion
$C(a)$ over $\Sigma$, if $\tau(a)\in \mathcal{H}$ and $\tau\in
T_\Sigma(C)$; and $\mathcal{F}$ \textit{satisfies} a role assertion
$P(a,b)$ (resp., $P^-(b,a)$) over $\Sigma$, if $P(a,b)\in
\mathcal{H}$. A feature $\mathcal{F}$ is a \textit{model feature} of KB $\mathcal{K}$ if $\mathcal{F}$ satisfies each
inclusion and each assertion in $\mathcal{K}$.
$Mod^F(\mathcal{K})$ denotes the set of all model features of
$\mathcal{K}$. It easily concludes that $\mathcal{K}$ is consistent
iff $Mod^F(\mathcal{K})\neq \emptyset$.
%
Given two KBs $\mathcal{K}_1$ and $\mathcal{K}_2$, let
$\Sigma=Sig(\mathcal{K}_1\cup \mathcal{K}_2)$,
$\mathcal{K}_1$ \textit{F-entails} $\mathcal{K}_2$ if
$Mod^F(\mathcal{K}_1)\subseteq Mod^F(\mathcal{K}_2)$, written by
$\mathcal{K}\models^F \mathcal{K}_2$; and $\mathcal{K}_1$ is
\textit{F-equivalent} $\mathcal{K}_2$ if $Mod^F(\mathcal{K}_1)=
Mod^F(\mathcal{K}_2)$, written by $\mathcal{K}\equiv^F
\mathcal{K}_2$.
In \cite{DBLP:conf/aaai/WangWT10}, we conclude that: \textit{(1)} $\mathcal{K}_1\models \mathcal{K}_2$ iff $\mathcal{K}_1\models^F \mathcal{K}_2$; \textit{(2)} $\mathcal{K}_1\equiv \mathcal{K}_2$ iff $\mathcal{K}_1\equiv^F \mathcal{K}_2$.
\section{Distance-based semantics for TBoxes}\label{sec:DT}
%

To measure the closeness of two types, we first define a distance function between two types in terms of the symmetric difference for sets.
\begin{definition}\label{def:distance function}
Let $\Sigma$ be a signature, a total function $d:T_{\Sigma}\times T_{\Sigma}\rightarrow \mathbb{R}^{+} \cup \{0\}$ is a \emph{pseudo-distance function} (for short, \emph{distance function}) on $T_{\Sigma}$ if it satisfies:(1) $\forall \tau_{1},\tau_{2}\in T_{\Sigma}, d(\tau_{1},\tau_{2})=0$ iff $\tau_{1}=\tau_{2}$; and (2) $\forall \tau_{1},\tau_{2}\in T_{\Sigma}, d(\tau_{1},\tau_{2})=d(\tau_{2},\tau_{1})$.
\end{definition}

Given a type $\tau\in T_{\Sigma}$ and a type set $\Xi\subseteq T_{\Sigma}$, the distance function between $\tau$ and $\Xi$ is defined as $d(\tau, \Xi)=min\{d(\tau,\tau')\mid \tau'\in \Xi\}$.

If $\Xi=\emptyset$, then we set $d(\tau, \Xi)=\mathbf{d}$ where $\mathbf{d}$ is a default value of distance function greater than any value be to considered. This setting is used to exclude all contradictions (e.g., $\top \sqsubseteq \bot$) under our candidate semantics since a contradiction can bring less useful information.


There are two representative distance functions on types, namely, \emph{Hamming distance function} where $d^{H}(\tau_{1},\tau_{2})=|(\tau_{1} - \tau_{2})\cup(\tau_{2} - \tau_{1})|$ and \emph{drastic distance function} where $d^{D}(\tau_{1},\tau_{2})=0$ if $\tau_{1}=\tau_{2}$ and $d^{D}(\tau_{1},\tau_{2})=1$ otherwise.

%
%
An \emph{aggregation function} $f$ is a total function that accepts
a multi-set of real numbers and returns a real number, satisfying:
(1) $f$ is non-decreasing in the values of its argument; (2) $f(\{x_1,\ldots,x_n\})=0$ iff $x_1=\ldots=x_n=0$; and (3) $\forall x\in \mathbb{R}^{+}\cup \{0\}$, $f(\{x\})=x$.
%
%

\begin{definition}\label{def:type-group}
Let $\Sigma$ be a signature, $\tau$ a type and $\Pi=\{\Xi_1,\ldots, \Xi_n\}$ a type group. Given a distance function $d$ and an aggregation function $f$, $\lambda_{d,f}$ between $\tau$ and $\Pi$ is defined as $\lambda_{d,f}(\tau,\Pi)=f(\{d(\tau,\Xi_1),\ldots,d(\tau,\Xi_n)\})$. Furthermore, $\tau$ is called $df$-\emph{minimal} (for short, \emph{minimal}) w.r.t. $\Pi$ if for any type $\tau'\in T_\Sigma$, $\lambda_{d,f}(\tau,\Pi)\leq\lambda_{d,f}(\tau',\Pi)$.
\end{definition}

We use $\Lambda_{d,f}(\Pi,\Xi)$ to denote a set of all $df$-minimal types w.r.t. $\Pi$ in $\Xi$.
%
\begin{proposition}\label{prop:inter}
Let $\Sigma$ be a finite signature and $\Pi=\{\Xi_1,\ldots, \Xi_n\}$ a type group over $\Sigma$. For any distance function $d$ and any aggregation function $f$, we have
(1) $\Lambda_{d,f}(\Pi,T_\Sigma)\neq\emptyset$ and (2) If $\cap \Pi\neq \emptyset$ then $\Lambda_{d,f}(\Pi,T_{\Sigma})=\cap \Pi$.
\end{proposition}

The first statement guarantees that a minimal type of it always exists if a type group contains a non-empty type set and the second shows that each type belong to all members of a type group is exactly a minimal type.

Let $\Sigma$ be a signature and $\mathcal{T}=\{\psi_1,\ldots,\psi_n\}$ a
TBox over $\Sigma$. Each axiom $\psi_i$ is of the form $C_i\sqsubseteq
D_i$ ($1\leq i \leq n$) where $C_i, D_i$ ($1\leq i \leq n $) are concepts.
We simply write $\Pi_{\Sigma}(\mathcal{T})$ as $\Pi(\mathcal{T})$ if $\Sigma=Sig(\mathcal{T})$.

\begin{corollary}\label{cor:p-consistent}
Let $\Sigma$ be a finite signature and $\mathcal{T}$ a TBox over $\Sigma$.
For any distance function $d$ and any aggregation function $f$, we have (1) $\Lambda_{d,f}(\Pi_{\Sigma}(\mathcal{T}),T_{\Sigma})\neq \emptyset$; and (2) If $\mathcal{T}$ is consistent then $\Lambda_{d,f}(\Pi_{\Sigma}(\mathcal{T}),T_{\Sigma})=\cap\Pi_{\Sigma}(\mathcal{T})$.
\end{corollary}
%

Unfortunately, $\Lambda_{d,f}(\Pi(\mathcal{T}),T_{\Sigma})$ does not always satisfy the role coherence as the following example shows.
\begin{example}\label{exe:3}
Let $\mathcal{T}=\{\top \sqsubseteq A\sqcap \exists P, \exists P^{-} \sqsubseteq \bot\}$ and $\Sigma=Sig(\mathcal{T})$. If $d$ is the Hamming distance function and $f$ is the summation function, then $\Lambda_{d^{H},f^{s}}(\Pi(\mathcal{T}),T_\Sigma)=\{\{A,\exists P\}\}$. Note that $\exists P^{-}\not \in \cup\Lambda_{d^{H},f^{s}}(\Pi(\mathcal{T}), T_\Sigma)$.
\end{example}

The reason that the role coherence might be absent in $\Lambda_{d,f}(\Pi(\mathcal{T}),T_{\Sigma})$ is that $\exists P$ and $\exists P^{-}$ are taken as two independent concepts so that the relation of satisfiability between $\exists P$ and $\exists P^{-}$ cannot be captured when minimal types are computed \cite{DBLP:conf/aaai/ZhuangWWQ14}.
%
To construct a model type set from a random type set $\Xi$, we introduce an iterative operator $\mu_{d,f}(\Xi)$ and its fixpoint.

Formally, let $\Sigma$ be a finite signature and $\Pi$ a type group over $\Sigma$. Given a type set $\Xi$ over $\Sigma$, let $\mu_{d,f}(\Xi)= \Xi \cup \Xi'$, where $\Xi' \subseteq T_{\Sigma}$ and $\Xi'= \{\tau\mid$ for some role $R$, $\exists R\in \cup\Xi$ and $\exists R^{-} \not\in \cup\Xi$, $\exists R^{-} \in \tau$ and for any type $\tau' \in T_{\Sigma}$, $\exists R^{-}\in \tau' $ implies $\lambda_{d,f}(\tau,\Pi)\leq \lambda_{d,f}(\tau',\Pi)\}$. We use $\Xi^{+}$ to denote the \textit{fixpoint} of $\mu_{d,f}$, i.e., $\Xi^{+} = FP(\mu_{d,f})(\Xi)$. For any distance function $d$, any aggregation function $f$, and any type set $\Xi$, we can conclude that $\Xi^{+}$ always exists since $\mu_{d,f}$ is inflationary (i.e., $\Xi \subseteq \mu_{d,f}(\Xi)$) and $\Sigma$ is finite.

Given a signature $\Sigma$ and a TBox $\mathcal{T}$ over $\Sigma$, we say $\Lambda^{+}_{d,f}(\Pi(\mathcal{T}),T_{\Sigma})$ is the \textit{minimal model type set} of $\mathcal{T}$. Intuitively, a minimal model type set is a set of minimal types with maintaining role coherence. In Example \ref{exe:3}, $\Lambda^{+}_{d^{H},f^{s}}(\Pi(\mathcal{T}),T_{\Sigma})$ = $\Lambda_{d^{H},f^{s}}(\Pi(\mathcal{T}),T_{\Sigma})\cup \{\tau_{3}\}$ = $\{\{A,\exists P\}, \{A,\exists P, \exists P^{-}\}\}$.

We show that minimal model type sets meet our motivation.
\begin{proposition}\label{prop:minimal-model-type}
Let $\Sigma$ be a signature and $\mathcal{T}$ a TBox over $\Sigma$. For any distance function $d$ and aggregation function $f$, we have
\begin{itemize}
\item $\Lambda^{+}_{d,f}(\Pi_{\Sigma}(\mathcal{T}),T_{\Sigma})\neq \emptyset$;
\item $\Lambda^{+}_{d,f}(\Pi_{\Sigma}(\mathcal{T}),T_{\Sigma})=\cap\Pi_{\Sigma}(\mathcal{T})$, if $\mathcal{T}$ is coherent;
\item $\exists P\in \cup \Lambda^{+}_{d,f}(\Pi_{\Sigma}(\mathcal{T}),T_{\Sigma})$ iff $\exists P^{-}\in \cup \Lambda^{+}_{d,f}(\Pi_{\Sigma}(\mathcal{T}),T_{\Sigma})$ for any $P\in \Sigma_{R}$.
\end{itemize}
\end{proposition}

In Proposition \ref{prop:minimal-model-type}, the first item states that there always exist minimal model types for any non-empty TBox; the second shows that when a TBox is consistent, each minimal model type is exactly model type; and the third ensures that minimal model type sets always satisfy the role coherence.
{proposition}
\begin{definition}\label{def:d-entail}
Let $\Sigma$ be a signature, $\mathcal{T}$ a TBox, and, $\phi$ an inclusion over $\Sigma$. Given a distance function $d$ and an aggregation function $f$, $\mathcal{T}$ distance-based entails (d-entails) $\phi$, denoted by $\mathcal{T}\models_{d,f}\psi$, if $\Lambda^{+}_{d,f}(\Pi_{\Sigma}(\mathcal{T}),T_{\Sigma})\subseteq Mod^T(\{\phi\})$.
\end{definition}

In Example \ref{exe:3}, $\mathcal{T}\models_{d^{H},f^{s}} \top \sqsubseteq A$. 

\section{Distance-based semantics for knowledge bases}\label{sec:DT-KB}
%
Compared with inconsistency of TBoxes, inconsistency occurring in KBs is much more complex.
For instance,
\begin{example}\label{exam:inconsistent}
Let $\mathcal{K}=(\{\exists P^-\sqsubseteq \bot\},\{\exists P(a)\})$ be a KB and $\Sigma=\{P,a, 1\}$. $\mathcal{K}$ is inconsistent and thus has no model feature.
\end{example}


We first introduce \textit{concept profiles} and then use type distance function to describe how far apart features are. Let $\Sigma$ be a signature and $\mathcal{A}$ an ABox over $\Sigma$. Assume that $N_A$ a set of all named individuals in $\mathcal{A}$.
$\mathcal{A}_R = \{P(a,b) \mid P(a,b)$ or $P^-(b,a)\in\mathcal{A}\}$. A \emph{concept profile} of $a$ in $\mathcal{A}$, denoted by $\Sigma_C(a)$, defined as follows: 
\begin{eqnarray*}
\Sigma_C(a) & = &  \bigcup_{D(a)\in \mathcal{A}}\{D\} \cup \bigcup_{P(a,b_1),\ldots,P(a,b_n)\in \mathcal{A}_R} \{\geq m\, P \mid m\in\Sigma_N, m\le n\} \\
& \cup & \bigcup_{P(b_1,a),\ldots,P(b_{n},a)\in\mathcal{A}_R} \{\geq m\, P^- \mid m\in\Sigma_N, m \le n \}.
\end{eqnarray*}


Let $\mathcal{K}=(\mathcal{T},\mathcal{A})$ be a KB. We extend the signature $Sig({K})$ of $\mathcal{K}$ as $Sig^{\ast}(\mathcal{K}) = Sig(\mathcal{T})\cup Sig(\Sigma_{C}(\mathcal{A}))$ where $\Sigma_{C}(\mathcal{A})=\bigcup_{a\in N_A} \Sigma_C(a)$. Indeed, $Sig^{\ast}(\mathcal{K})$ is obtained from $Sig(\mathcal{K})$ by adding all possible natural numbers occurring all concept profiles but not occurring $\mathcal{K}$.

Next, we will define the notion of minimal model features.
\begin{definition}\label{def:mini-feature}
Let $\Sigma$ be a signature and $\mathcal{K}=(\mathcal{T},\mathcal{A})$ a KB over $\Sigma$. Denote $\Pi_{\Sigma}(a)=\{T_\Sigma (D)\mid D\in \Sigma_C(a)\}$. Given a distance function $d$ and an aggregation function $f$, a $df$-\emph{minimal model feature} of $\mathcal{K}$ is a feature $\mathcal{F}=\langle \Xi, \mathcal{H}\rangle$ satisfying the following four conditions:
\begin{itemize}
  \item $\Xi\subseteq \Lambda^{+}_{d,f}(\Pi_{\Sigma}(\mathcal{T}), T_{\Sigma})$;
  \item for each $P\in \Sigma_R$, $\exists P\in \cup \Xi$ iff $\exists P^-\in
  \cup\Xi$;
  \item $\tau\in\Lambda^{+}_{d,f}(\Pi_{\Sigma}(a), \Lambda^{+}_{d,f}(\Pi_{\Sigma}(\mathcal{T}), T_{\Sigma}))\cap\Xi$ for each $a \in \Sigma_I$ and $\tau(a) \in\mathcal{H}$;
  \item for any role assertion $P(a,b)\in\mathcal{A}_R - \mathcal{H}$, either
$\ge n+1\, P(a)\not\in\mathcal{H}$ and $P(a,b_1)$, $\ldots$, $P(a,b_{n})\in
\mathcal{H}$, or $\ge n+1\, P^-(b)\not\in\mathcal{H}$ and $P(a_1,b), \ldots, P(a_{n},b) \in \mathcal{H}$.
\end{itemize}
$Mod^F_{d,f}(\mathcal{K})$ is the set of $df$-minimal model features of $\mathcal{K}$.
\end{definition}

In Definition \ref{def:mini-feature}, 
the first condition requires that all types of $\Xi$ be minimal; the second says that $\Xi$ should be a model type set; the third guarantees that each type of $\Xi$ satisfying each concept assertion in $\mathcal{H}$ has the minimal distance function to its corresponding concept profile; and the last ensures that $\mathcal{F}$ is consistent by those role assertions conflicting with concept assertions.
\begin{example}\label{exam:Penguin-KB}
In \textit{Penguin} KB, we abbreviate $\textit{Penguin}$ to $P$, $\textit{Swallow}$ to $S$, $\textit{Bird}$ to $B$, $\textit{Fly}$ to $F$, $\textit{tweety}$ to $t$ and $\textit{fred}$ to $r$. Let $\Sigma=\{P,S,B,F,t,r\}$, $\Sigma_C(t)=\{P, \neg F\}$ and $\Sigma_C(r)=\{S\}$. Assume that $d$ is the Hamming distance function and $f$ is the summation function. We have $\Lambda^{+}_{d,f}(\Pi_{\Sigma}(\mathcal{T}), T_{\Sigma})$ $=\{\tau_1,\tau_2,\tau_4,\tau_8,\tau_{12},\tau_{16}\}$. Here  $\tau_{1}=\{\}$, $\tau_{2}=\{F\}$, $\tau_{4}=\{B,F\}$,$\tau_{8}=\{S,B,F\}$, $\tau_{12}=\{P,B,F\}$, and $\tau_{16}=\{P,S,B,F\}$.all of whose distance is 0.
%
We have that
$\Lambda^{+}_{d,f}(\Pi_\Sigma(t),\Lambda^{+}_{d,f}(\Pi_{\Sigma}(\mathcal{T}), T_{\Sigma}))$ = $\{\tau_1,\tau_{12},\tau_{16}\}$ and $\Lambda^{+}_{d,f}(\Pi_\Sigma(r),\Lambda^{+}_{d,f}(\Pi_{\Sigma}(\mathcal{T}), T_{\Sigma}))$ = $\{\tau_8, \tau_{16}\}$.
Note that all types in $\Lambda^{+}_{d^{H},f^{s}}(\Pi_\Sigma(t),T_{\Sigma})$ have distance equal to $1$ while all types in $\Lambda^{+}_{d^{H},f^{s}}(\Pi_\Sigma(r),T_{\Sigma})$ have distance equal to $0$.
Thus, $Mod^F_{d,f}(\mathcal{K})=$ $\{\langle \Xi, \tau(t)\cup \tau'(r)\rangle\mid$ $\tau\in\{\tau_1,\tau_{12},\tau_{16}\}$, $\tau'\in \{\tau_8,\tau_{16}\}$, $\{\tau,\tau'\}\subseteq \Xi$ and $\Xi\subseteq \{\tau_1,\tau_8,\tau_{12},\tau_{16}\}\}$.
\end{example}

We find that minimal model features can reach our aim.
\begin{proposition}\label{prop:minimal-model-fearture}
Let $\Sigma$ be a signature and $\mathcal{K}$ a KB over $\Sigma$. For any distance function $d$ and any aggregation function $f$, we have
\begin{itemize}
\item $Mod^F_{d,f}(\mathcal{K}) \neq \emptyset$;
\item $Mod^F_{d,f}(\mathcal{K})=Mod^F(\mathcal{K})$, if $\mathcal{K}$ is consistent.
\end{itemize}
\end{proposition}
An expected result is that the second statement of Proposition \ref{prop:minimal-model-fearture} does not necessarily hold if $\mathcal{K}$ is inconsistent. For instance, in Example \ref{exam:inconsistent}, $Mod^F_{d,f}(\mathcal{K})=\{\mathcal{F}_{1}, \mathcal{F}_{2}\}$ where $\mathcal{F}_{1}=\langle \{\exists P\}, \{\exists P(a)\}\rangle$ and $\mathcal{F}_{2}=\langle \{\exists P,\exists P^{-}\}, \{\exists P(a),\exists P^{-}(a)\}\rangle$ while $Mod^F(\mathcal{K})=\emptyset$.

Now, based on minimal model features, we are ready to define the \emph{distance-based entailment} for KBs, written $\models_{d,f}$, under which meaningful information can be entailed from an inconsistent KB.

\begin{definition}\label{def:d-entail}
Let $\Sigma$ be a signature, $\mathcal{K}$ a KB, and, $\phi$ an axiom over $\Sigma$. Given a distance function $d$ and an aggregation function $f$, $\mathcal{K}$ distance-based entails (d-entails) $\phi$, still denoted by $\mathcal{K}\models_{d,f}\phi$, if $Mod^F_{d,f}(\mathcal{K})\subseteq Mod^F(\{\phi\})$.
\end{definition}

Distance-based entailment brings a new semantics (called \emph{distance-based semantics}) for inconsistent KBs by weakening classical entailment. It is not hard to see that no contradiction can be entailed in this semantics. For instance, in \textit{Penguin} KB, $\neg \textit{Fly}\sqcap \textit{Fly} (\textit{tweety})$ cannot be entailed but $\neg \textit{Fly}\sqcup \textit{Fly}(\textit{tweety})$ can under our semantics.

In the rest of this section, we exemplify that distance-based semantics is suitable for reasoning with inconsistent KBs.

Consequences are intuitive and reasonable under the distance-based semantics.
In \textit{Penguin} KB, $\mathcal{K}\models_{d^{H},f^{s}} \textit{Fly}(\textit{fred})$ while $\mathcal{K}\not\models_{d^{H},f^{s}} \textit{Penguin}(\textit{tweety})$ and $\mathcal{K}\not\models_{d^{H},f^{s}} \textit{Fly}(\textit{tweety})$. We further analyze those conclusions under distance-based semantics. The inconsistency of $\mathcal{K}$ is caused by statement about $\textit{tweety}$. On the one hand, $\textit{tweety}$ is a penguin which cannot fly, i.e., $\neg \textit{Fly}(\textit{tweety})$. On the other hand, a penguin is a bird which can fly, i.e., $\textit{Fly}(\textit{tweety})$. Moreover, there exists no more argument for either $\textit{Penguin}(\textit{tweety})$ or $\textit{Fly}(\textit{tweety})$. In this sense, neither $\textit{Penguin}(\textit{tweety})$ nor $\textit{Fly}(\textit{tweety})$ can be entailed under distance-based semantics. However, the statement about $\textit{fred}$ in $\mathcal{K}$ contains no conflict. Thus $\textit{Fly}(\textit{fred})$ can be entailed under distance-based semantics. Additionally, let us consider a simple example: let $\mathcal{A} = \{A(a), \neg A(a), B(b)\}$. We can conclude that $\mathcal{A} \models_{d^H, f^s} B(b)$ while neither $\mathcal{A} \not \models_{d^H, f^s} A(a)$ nor $\mathcal{A} \not \models_{d^H, f^s} A(a)$.
%

\section{Properties of distance-based semantics}\label{sec:DTS}
In this section, we present some useful properties of distance-based semantics. 

If $\mathcal{K}$ is inconsistent and there exists an axiom $\phi$ such that $\mathcal{K}\not\models_{p} \phi$ where $\models_{p}$ is an entailment relation, then we say $\models_{p}$ is \emph{paraconsistent}. It is well known that classical entailment $\models$ is not paraconsistent. We reconsider Example \ref{exam:inconsistent} and we have $\mathcal{K}\models_{d^{H},f^{s}} \exists P^-\sqsubseteq \bot$ while $\mathcal{K}\not\models_{d^{H},f^{s}}  \exists P(a)$. 

The following result shows that the distance-based entailment is paraconsistent.
\begin{proposition}\label{prop:paraconsistent}
For any distance function $d$ and any aggregation function $f$, $\models_{d,f}$ is paraconsistent.
\end{proposition}

Most existing semantics for paraconsistent reasoning in DLs are much weaker than the classical semantics in this sense that there exists a consistent KB $\mathcal{K}$ and an axiom $\phi$ such that $\mathcal{K}\models \phi$ (also called \emph{consistency preservation}) but $\phi$ is not entailed by $\mathcal{K}$ under the paraconsistent semantics. The following result shows that the distance-based semantics does not have such shortcoming.

We can conclude a result directly following Proposition \ref{prop:minimal-model-fearture}.
\begin{proposition}\label{prop:Consistent-P}
Let $\Sigma$ be a signature, $\mathcal{K}$ a KB, and, $\phi$ an axiom over $\Sigma$. For any distance function $d$ and any aggregation function $f$, if $\mathcal{K}$ is consistent then $\mathcal{K}\models_{d,f}\phi$ iff $\mathcal{K}\models \phi$.
\end{proposition}

In the classical semantics, a property that $\mathcal{K}\models \psi$ iff $\mathcal{T}\models \psi$ for any inclusion $\psi$ is called \emph{TBox-preservation} where the problem of subsumption checking is irrelevant to ABoxes. Our distance-based semantics satisfies such a property.
\begin{proposition}\label{prop:Preservation-TBox}
Let $\Sigma$ be a signature, $\mathcal{K}=(\mathcal{T},\mathcal{A})$ a KB, and, $\psi$ an inclusion over $\Sigma$. For any distance function $d$ and any aggregation function $f$, $\mathcal{K}\models_{d,f} \psi$ iff $\mathcal{T}\models_{d,f} \psi$.
\end{proposition}

By Proposition \ref{prop:Preservation-TBox}, TBox preservation property means that if the TBox by itself is consistent, then it will be entailed (and hence preference is given to preserving TBox statements over ABox statements), such as the same treatment in  \cite{DBLP:conf/rr/LemboLRRS10}. This is different from some other approaches to inconsistency-handling in DLs, where the TBox and ABox are treated equally, or the ABox is given preference such as  \cite{DBLP:conf/dlog/MaP10,DBLP:journals/jiis/ZhangL13,DBLP:journals/ijar/ZhangXLJ14}.

The closure w.r.t. $\models_{d,f}$ of an arbitrary KB is always consistent.
\begin{proposition}\label{prop:d-consistent}
Let $\Sigma$ be a signature and $\mathcal{K}=(\mathcal{T},\mathcal{A})$ a KB over $\Sigma$. For any distance function $d$ and any aggregation function $f$, let $Cn_{d,f}(\mathcal{T}) = \{ \psi$ is an inclusion $\mid \mathcal{T}\models_{d,f} \psi \}$ and $Cn^{\mathcal{T}}_{d,f}(\mathcal{A}) = \{ \varphi$ is an assertion $\mid (\mathcal{T},\mathcal{A})\models_{d,f} \varphi \}$. We conclude that both $Cn_{d,f}(\mathcal{T})$ and $Cn^{\mathcal{T}}_{d,f}(\mathcal{A})$ are
consistent.
\end{proposition}

Proposition \ref{prop:d-consistent} provides a theoretical foundation of applying our approach to \emph{inconsistency}-\emph{tolerant conjunctive query answering} \cite{DBLP:conf/ijcai/BienvenuR13}. 

Let $\Sigma$ be a signature. A distance function $d$ is $\Sigma$-\emph{unbiased}, if for any $\Sigma$-concept $C$ and any two $\Sigma$-types $\tau_{1},\tau_{2}$, $B\in \tau_{1}$ iff $B\in \tau_{2}$ for any basic concept $B$ occurring in $C$ implies $d(\tau_{1},T_{\Sigma}(C))=d(\tau_{2},T_{\Sigma}(C))$. The Hamming distance function and the drastic distance function are unbiased. 

Let us consider a distance function $d^{\cup}$ defined as follows: for any two sets $S_1, S_2$, $d^{\cup}(S_1, S_2) = 0$ if $S_1 = S_2$; and $d^{\cup}(S_1, S_2) = 1+ |S_1 \cup S_2|$. It clearly concludes that $d^{\cup}(S_1, S_2) = 0$ iff $S_1 = S_2$ and $d^{\cup}(S_1, S_2) = d^{\cup}(S_2, S_1) $. Thus $d^{\cup}$ is a distance function.
Let $\Sigma = \{A_1, A_2, A_3, A_4\}$ and $C = A_1 \sqcap A_2$. For each type $\tau \in T_{\Sigma}(C)$, $\{A_1, A_2\} \subseteq \tau$. Let $\tau_1 = \{A_1, A_2\}$ and $\tau_2 = \{A_1, A_2, A_3, A_4\}$. Thus $d(\tau_{1},T_{\Sigma}(C)) = 5$ and $d(\tau_{2},T_{\Sigma}(C)) = 7$. Then $d^{\cup}$ is not unbiased.

Unbiasedness will bring a good property of relevance in reasoning.
\begin{proposition}\label{prop:relevance}
Let $\Sigma$ be a signature, $\mathcal{K}$ a KB, and, $\phi$ a non-tautology over $\Sigma$. If $d$ is an unbiased function and $Sig(\mathcal{K})\cap Sig(\{\phi\})=\emptyset$ then for any aggregation function $f$, $\mathcal{K}\not\models_{d,f} \phi$.
\end{proposition}

Note that Proposition \ref{prop:relevance} does no longer true for tautologies. Let $\mathcal{K}$ be a KB and $\phi$ a tautology with $Sig(\mathcal{K}) \cap Sig(\phi) = \emptyset$, let $\Sigma = Sig(\mathcal{K}) \cup Sig(\phi)$, we can conclude that for any distance function $d$ and any aggregation function $f$, $\mathcal{K} \models_{d,f}\phi$ since all possible $\Sigma$-features can satisfy $\phi$. 

An entailment relation $\models_{m}$ is \emph{monotonic} if $\mathcal{K}' \models_{m} \phi$ implies $\mathcal{K}\models_{m} \phi$ for any KB $\mathcal{K}'\subseteq \mathcal{K}$; and \emph{nonmonotonic} otherwise. Another characteristic property of $\models_{d,f}$ is its non-monotonic nature.
\begin{proposition}\label{prop:nonmonotonic}
For any distance function $d$ and any aggregation function $f$, $\models_{d,f}$ is non-monotonic.
\end{proposition}

While the distance-based semantics is non-monotonic in general, it
satisfies a kind of cautious monotonicity, which is usually referred
to as \emph{splitting property}.


We say $\mathcal{K}$ is split into $\mathcal{K}'$ and
$\mathcal{K}''$, denoted $\mathcal{K}=\mathcal{K}'\oplus
\mathcal{K}''$, if (1) $\mathcal{K}=\mathcal{K}'\cup \mathcal{K}''$,
and (2) $Sig(\mathcal{K}')\cap Sig(\mathcal{K}'')=\emptyset$.

An aggregation function $f$ is \emph{hereditary} iff
$f(\{x_1,\ldots,x_n\})$ $<f(\{y_1,\ldots,y_n\})$ implies for any
$z_1,\ldots,z_m$, $f(\{x_1,\ldots,x_n,z_1,\ldots,z_m\})$ $<f(\{y_1,\ldots,y_n, z_1,\ldots,z_m\})$. 

\begin{proposition}\label{prop:partition}
Let $\Sigma$ be a signature and $\mathcal{K}$ a KB over $\Sigma$. Assume that $\mathcal{K}=\mathcal{K}'\oplus \mathcal{K}''$ where $\mathcal{K}'$ is consistent. For each axiom $\phi$ with $Sig(\phi)\cap Sig(\mathcal{K}'')=\emptyset$, if $\mathcal{K}'\models \phi$ then for any distance function $d$ and any hereditary aggregation function $f$, $\mathcal{K}\models_{d,f}\phi$.
\end{proposition}

One advantage of the splitting property is that the paraconsistent
reasoning in KB $\mathcal{K}$ can be localized into the
classical reasoning in a consistent module of $\mathcal{K}$, which
is usually smaller than the original $\mathcal{K}$. Such a
property can be very useful for a highly distributed ontology
system.

A relation $|{\approx}$ is \emph{cautious} if it satisfies:
\begin{itemize}
\item (\emph{cautious reflexivity}) If $\mathcal{K}=\mathcal{K}'\oplus \mathcal{K}''$ and $\mathcal{K}'$ is consistent, then $\mathcal{K}|{\approx}\varphi$ for all axiom $\varphi\in \mathcal{K}'$;
\item (\emph{cautious monotonicity}) If $\mathcal{K}|{\approx} \varphi$ and $\mathcal{K}|{\approx} \psi$, then $\mathcal{K}\cup \{\varphi\}|{\approx} \psi$;
\item (\emph{cautious cut}) If $\mathcal{K}|{\approx} \varphi$ and  $\mathcal{K}\cup \{\varphi\}|{\approx} \psi$ then $\mathcal{K}|{\approx} \psi$.
\end{itemize}

\begin{proposition}\label{prop:cautious}
For any distance function $d$ and any monotonic hereditary aggregation function $f$, $\models_{d,f}$ is cautious.
\end{proposition}
\begin{example}\label{exam:haswife}
Consider an ABox $\mathcal{A}=$ $\{$ $\textit{HasWife}(\textit{Mike},\textit{Rose})$, $\textit{HasWife}(\textit{Mike}, \textit{Mary})\}$. Let $\Sigma = \{\textit{HasWife}, \textit{Mike}, \textit{Mary}, \textit{Rose}, 1, 2\}$. The first statement claims that $\textit{Mike}$ has at most one wife. Moreover, we are informed that $\textit{Mike}$ has two wives $\textit{Rose}$ and $\textit{Mary}$. We conclude that $\mathcal{A}$ is inconsistent and $\mathcal{A}\models_{d^{H},f^{s}} \ge\, 1 \textit{HasWife}(\textit{Mike})$ while $\mathcal{A}\not\models_{d^{H},f^{s}} \textit{HasWife}(\textit{Mike}, \textit{Rose}),$ and
$\mathcal{A}\not\models_{d^{H},f^{s}}$ $\textit{HasWife}(\textit{Mike}, \textit{Mary})$.
Intuitively, $\textit{Mike}$ has a wife while we don't know whether his wife is $\textit{Rose}$ or $\textit{Mary}$ under our distance-based semantics.
\end{example}

\section{Discussions}\label{sec:related}
In this paper, we have presented a model-based framework to handle inconsistency in DL-Lite by introducing distances over types of features for KBs. Within this framework, we defined a new semantics called distance-based semantics. Furthermore, our framework gives consideration to both semantic minimal change and syntactic minimal change. In this sense, our approach is a natural combination of qualitative and quantitative approaches. 

Existing model-centered approaches for inconsistency handling are usually based on various forms of inconsistency-tolerant semantics, such as four-valued description logics \cite{DBLP:conf/rr/MaH09,DBLP:conf/dlog/MaP10}, quasi-classical description logics \cite{DBLP:journals/ijar/ZhangXLJ14}, argumentation-based semantics for description logics \cite{DBLP:journals/logcom/Kamide12,DBLP:journals/jiis/ZhangL13}, and the MKNF-based semantics for description logics \cite{DBLP:journals/igpl/HuangLH13}. Compared to them, our distance-based semantics works on classical interpretations but still can draw more useful and reasonable logical consequences. 
Moreover, these approaches do not provide a mechanism of comparing different models for a KB and are usually monotonic such that they do not hold consistency-preserving. 
The argumentation-based semantics for description logics presented in \cite{DBLP:journals/logcom/Kamide12,DBLP:journals/jiis/ZhangL13} is based on a dialogue process to evaluate the inconsistent knowledge. Our semantics is based on a totally different mechanism from it. Different from \cite{DBLP:journals/igpl/HuangLH13} which introduces a weak negation \textbf{not} to tolerate inconsistency, our approach does not change the syntax of DLs.
Different from syntax-based paraconsistent approaches taking some consistent subsets as substitutes of KBs in reasoning \cite{DBLP:conf/ijcai/SchlobachC03,DBLP:conf/ijcai/HuangHT05,DBLP:conf/esws/KalyanpurPSG06,DBLP:conf/aaai/MeyerLBP06,DBLP:conf/www/DuS08,DBLP:conf/semweb/RosatiRGM11}. Similarly to our approach, those syntax-based paraconsistent semantics can satisfy several properties that do not hold in multi-valued semantics, such as non-monotonicity, consistency-preserving and splitting property. But they differ from ours in the following aspects. Firstly, they do not satisfy the closure consistency. Secondly, those syntax-based approaches focus on local information so that they could difficultly capture the semantics of whole a KB. Finally, they might bring the multi-extension problem because of limitations of their selection mechanisms.

There are some model-based approaches presented in \cite{DBLP:conf/ijcai/QiD09,DBLP:conf/rr/LemboLRRS10}. A model-repaired approach is presented to recover the consistency arising from adding one TBox to the other one \cite{DBLP:conf/ijcai/QiD09}. Compared with it directly working on models, our approach works on types which take take advantage of finiteness. 
%
Some repairing approaches are applied to repair ABoxes such that the repaired KB can ensure the union of conjunctive consistent querying when ABoxes conflict with TBoxes \cite{DBLP:conf/rr/LemboLRRS10}. Although both of the main goal of this work and our work are recovering consistency by repairing KBs, there exists some difference in strategies. They repair an ABox according to a consistent TBox. However, we construct those models which are closer to a KB according to some distance function and aggregation function when there exists no model in an inconsistent KB.

A distance-based approach is proposed to measure inconsistency of TBoxes \cite{DBLP:conf/dlog/MaP10}. However, this approach might be difficult to do so because of infinite number of models of DL KBs since it is based on the distance between models. As a future work, we employ our distance-based technique to measure inconsistency of KBs. 



\newpage
\section*{Appendix: proofs}

\noindent \textbf{Proposition} \ref{prop:inter}\\
Let $\Sigma$ be a finite signature and $\Pi=\{\Xi_1,\ldots, \Xi_n\}$ a type group over $\Sigma$. For any distance function $d$ and any aggregation function $f$, we have
\begin{itemize}
  \item $\Lambda_{d,f}(\Pi,T_\Sigma)\neq\emptyset$;
  \item If $\cap \Pi\neq \emptyset$ then $\Lambda_{d,f}(\Pi,T_{\Sigma})=\cap \Pi$.
\end{itemize}
\begin{proof}
\begin{itemize}
\item For any type $\tau\in T_{\Sigma}$, we can compute that $\lambda_{d,f}(\tau,\Pi)$ by Definition \ref{def:type-group}. That is, all $\lambda_{d,f}(\tau,\Pi)$ are comparable. 
Suppose, for the sake of contradiction, that $\Lambda_{d,f}(\Pi,T_\Sigma) = \emptyset$. Then, by the definition of $\Lambda_{d,f}(\Pi,T_\Sigma)$, for any $\tau \in T_\Sigma$, there exists some $\tau' \in T_\Sigma$ such that $\lambda_{d,f}(\tau',\Pi) < \lambda_{d,f}(\tau,\Pi)$. Thus $T_\Sigma$ is infinite. However, $T_\Sigma$ is finite since $\Sigma$ is finite, we have arrived at a contradiction. 
Therefore, $\Lambda_{d,f}(\Pi,T_{\Sigma})\neq \emptyset$ by the definition of distance functions and aggregation functions.
\item On the one hand, if $\cap \Pi \neq \emptyset$ then for any type $\tau\in \cap \Pi$, $\lambda_{d,f} (\tau,\Pi)\leq  \lambda_{d,f}(\tau',\Pi)$ for any type $\tau'\in T_{\Sigma}$ since $\tau \in \cap \Pi$, i.e., $\lambda_{d,f}(\tau,\Pi)=0$ by Definition \ref{def:distance function} and Definition \ref{def:type-group}. On the other hand, if $\tau\in \Lambda_{d,f}(\Pi,T_{\Sigma})$, then for any type $\tau''\in T_{\Sigma}$, $\lambda_{d,f} (\tau,\Pi)\leq  \lambda_{d,f}(\tau'',\Pi)$ by the definition. We choose $\tau''\in \cap\Pi \subseteq T_{\Sigma}$. So $\lambda_{d,f} (\tau,\Pi)=0$ since $\lambda_{d,f} (\tau'',\Pi)=0$ by the proof if the first item.
\end{itemize}
\end{proof}

\noindent \textbf{Corollary} \ref{cor:p-consistent}\\
Let $\Sigma$ be a finite signature and $\mathcal{T}$ a TBox over $\Sigma$. For any distance function $d$ and any aggregation function $f$, we have
\begin{itemize}
\item $\Lambda_{d,f}(\Pi_{\Sigma}(\mathcal{T}),T_{\Sigma})\neq \emptyset$;
\item if $\mathcal{T}$ is consistent then $\Lambda_{d,f}(\Pi_{\Sigma}(\mathcal{T}),T_{\Sigma})=\cap\Pi_{\Sigma}(\mathcal{T})$.
\end{itemize}
\begin{proof}
The first item directly follows the first item of Proposition \ref{prop:inter}. In the second item, if $\mathcal{T}$ is consistent then $\cap \Pi_\Sigma(\mathcal{T}) \neq \emptyset$. By the second item of Proposition \ref{prop:inter}, we can conclude that $\Lambda_{d,f}(\Pi_{\Sigma}(\mathcal{T}),T_{\Sigma})=\cap\Pi_{\Sigma}(\mathcal{T})$.
\end{proof}

\noindent \textbf{Proposition} \ref{prop:minimal-model-type}\\
Let $\Sigma$ be a signature and $\mathcal{T}$ a TBox over $\Sigma$. For any distance function $d$ and aggregation function $f$, we have
\begin{itemize}
\item $\Lambda^{+}_{d,f}(\Pi_{\Sigma}(\mathcal{T}),T_{\Sigma})\neq \emptyset$;
\item $\Lambda^{+}_{d,f}(\Pi_{\Sigma}(\mathcal{T}),T_{\Sigma})=\cap\Pi_{\Sigma}(\mathcal{T})$, if $\mathcal{T}$ is coherent;
\item $\exists P\in \cup \Lambda^{+}_{d,f}(\Pi_{\Sigma}(\mathcal{T}),T_{\Sigma})$ if and only if $\exists P^{-}\in \cup \Lambda^{+}_{d,f}(\Pi_{\Sigma}(\mathcal{T}),T_{\Sigma})$ for any role name $P\in \Sigma_{R}$.
\end{itemize}
\begin{proof}
We can use Corollary \ref{cor:p-consistent} and the definitions to prove this proposition.
\begin{itemize}
\item The first item directly follows the first item of Corollary \ref{cor:p-consistent} and the definition of minimal model type sets since $\Lambda_{d,f}(\Pi_{\Sigma}(\mathcal{T}),T_{\Sigma}) \subseteq \Lambda^{+}_{d,f}(\Pi_{\Sigma}(\mathcal{T}),T_{\Sigma})$ and \\ $\Lambda_{d,f}(\Pi_{\Sigma}(\mathcal{T}),T_{\Sigma}) \neq \emptyset$.
\item In the second item, if $\mathcal{T}$ is consistent then $\Lambda_{d,f}(\Pi_{\Sigma}(\mathcal{T}),T_{\Sigma})=\cap \Pi_{\Sigma}(\mathcal{T})$ by the second of Corollary \ref{cor:p-consistent}. Because $\cap \Pi_{\Sigma}(\mathcal{T})$ is the model type set of $\mathcal{T}$, $\exists P \in \cup(\cap \Pi_{\Sigma}(\mathcal{T}))$ if and only if $\exists P^{-} \in \cup(\cap \Pi_{\Sigma}(\mathcal{T}))$ for any role $P\in \Sigma_{R}$ by the definition of model type sets. Then for any role name $P\in \Sigma_{R}$, $\exists P\in \cup\Lambda_{d,f}(\Pi_{\Sigma}(\mathcal{T}),T_{\Sigma})$ if and only if  $\exists P^{-}\in \cup\Lambda_{d,f}(\Pi_{\Sigma}(\mathcal{T}),T_{\Sigma})$.\\ Therefore, $\Lambda^{+}_{d,f}(\Pi_{\Sigma}(\mathcal{T}),T_{\Sigma})=\Lambda_{d,f}(\Pi_{\Sigma}(\mathcal{T}),T_{\Sigma})$. That is, $\Lambda^{+}_{d,f}(\Pi_{\Sigma}(\mathcal{T}),T_{\Sigma})=\cap \Pi_{\Sigma}(\mathcal{T})$.
\item 
It directly follows the definition of $\Lambda^{+}_{d,f}(\Pi_{\Sigma}(\mathcal{T}),T_{\Sigma})$.
\end{itemize}
\end{proof}

\noindent \textbf{Proposition} \ref{prop:minimal-model-fearture}\\
Let $\Sigma$ be a signature and $\mathcal{K}$ a KB over $\Sigma$. For any distance function $d$ and any aggregation function $f$, we have
\begin{itemize}
\item $Mod^F_{d,f}(\mathcal{K}) \neq \emptyset$;
\item $Mod^F_{d,f}(\mathcal{K})=Mod^F(\mathcal{K})$, if $\mathcal{K}$ is consistent.
\end{itemize}
\begin{proof}
The first item directly follows the first item of Proposition \ref{prop:minimal-model-type} and Definition \ref{def:mini-feature}. 

In the second item, if $\mathcal{K}=(\mathcal{T},\mathcal{A})$ is consistent, then, $Mod^{F}(\mathcal{K})\neq \emptyset$.
\begin{enumerate}
\item 
For any feature $\mathcal{F}=\langle \Xi, \mathcal{H}\rangle\in Mod^{F}(\mathcal{K})$, for any feature $\mathcal{F}'=\langle \Xi',\mathcal{H}'\rangle$, we have
\begin{enumerate}
\item for any type $\tau\in\Xi$ and for any type $\tau'\in \Xi'$, $0=\lambda_{d,f}(\tau,\Pi_{\Sigma}(\mathcal{T}))\leq \lambda_{d,f}(\tau',\Pi_{\Sigma}(\mathcal{T}))$, that is, $\Xi \subseteq \Xi_{\mathcal{K}}$;
\item $\exists P \in \cup \Xi$ if and only if $\exists P^{-} \in \cup \Xi$;
\item for each individual $a\in \Sigma_{I}$ and $\tau(a)\in \mathcal{H}$ for some $\tau \in \Xi$, for any type $\tau'\in \Xi'$ and $\tau'(a)\in \mathcal{H}'$, $0=\lambda_{d,f}(\tau,T^{a}_{\Sigma})\leq \lambda_{d,f}(\tau',T^{a}_{\Sigma})$, that is, $\tau \in \Lambda_{d,f}(\tau,T^{a}_{\Sigma})$;
\item for each assertion $P(a,b)\in\mathcal{A}_R - \mathcal{H}$, either
$\le n P(a)\in\mathcal{H}$ and $P(a,b_1)$, $\ldots$, $P(a,b_{n+1})\in
\mathcal{A}_R$, or $\le n P^-(b) \in\mathcal{H}$ and
$P(a_1,b)$, $\ldots$, $P(a_{n+1},b) \in \mathcal{A}_R$.
\end{enumerate}
Based on (a), (b), (c) and (d), $\mathcal{F}\in Mod^{F}_{d,f}(\mathcal{K})$. \\
\item 
For any feature $\mathcal{F}=\langle \Xi, \mathcal{H}\rangle\in Mod^{F}_{d,f}(\mathcal{K})$ and $\mathcal{F}'=\langle \Xi',\mathcal{H}'\rangle\in Mod^{F}(\mathcal{K})$ since $Mod^{F}(\mathcal{K})\neq \emptyset$, we have
\begin{enumerate}
\item for any type $\tau\in \Xi$ and for any type $\tau'\in \Xi'$, $\lambda_{d,f}(\tau,\Pi_{\Sigma}(\mathcal{T}))\leq$ $ \lambda_{d,f}(\tau',\Pi_{\Sigma}(\mathcal{T}))=0$, that is, $\lambda_{d,f}(\tau,\Pi_{\Sigma}(\mathcal{T}))=0$. Then $\tau\in \cap \Pi_{\Sigma}(\mathcal{T})$, i.e., $\Xi \subseteq \cap \Pi_{\Sigma}(\mathcal{T})$. Therefore, $\mathcal{F}$ satisfies all inclusions in the TBox of $\mathcal{K}$.
\item $\exists P \in \cup \Xi$ if and only if $\exists P^{-} \in \cup \Xi$;
\item for each assertion $C(a)\in \mathcal{A}$, there exists $\tau' \in \Xi'$ such that $\tau' \in T^{a}_{\Sigma})$ and $\tau'(a)\in \mathcal{H}'$. For any type $\tau\in \Xi$, $\lambda_{d,f}(\tau,T^{a}_{\Sigma})\leq \lambda_{d,f}(\tau',T^{a}_{\Sigma})=0$, that is,
$\lambda_{d,f}(\tau,T^{a}_{\Sigma})=0$. Therefore, $\tau(a)\in \mathcal{H}$ and $\tau\in T_{\Sigma}(C)$. That is, $\mathcal{F}$ satisfies $C(a)$.
\item for each $P(a,b)\in \mathcal{A}$, $\mathcal{F}$ satisfies both $\exists P(a)$ and $\exists P^{-}(b)$ by the analogous proof of Item (b). $P(a,b)\in \mathcal{H}$ by the Item 4 of Definition \ref{def:mini-feature}.
\end{enumerate}
Based on (a), (b), (c), and (d), $\mathcal{F}\in Mod^{F}(\mathcal{K})$. Therefore, $Mod^{F}(\mathcal{K})=Mod^{F}_{d,f}(\mathcal{K})$.
\end{enumerate}
\end{proof}

\noindent {Proposition} \ref{prop:paraconsistent}\\
For any distance function $d$ and any aggregation function $f$, $\models_{d,f}$ is paraconsistent.\\
\begin{proof}
Let $\Sigma$ be a signature. Let $\mathcal{K}$ be a KB over $\Sigma$. For any contradiction $\phi$, for any distance function $d$ and for any aggregation function $f$, we can conclude that $Mod^F_{d,f}(\mathcal{K}) \neq \emptyset$ by the first item of Proposition \ref{prop:minimal-model-fearture}. Because $\phi$ is a contradiction, $Mod^F(\{\phi\}) = \emptyset$. Therefore, $Mod^F_{d,f}(\mathcal{K}) \not \subseteq Mod^F(\{\phi\})$, that is, $\mathcal{K} \not\models_{d,f} \phi$.
\end{proof}

\noindent \textbf{Proposition}\ref{prop:Preservation-TBox}\\
Let $\Sigma$ be a signature, $\mathcal{K}=(\mathcal{T},\mathcal{A})$ a KB, and, $\psi$ an inclusion over $\Sigma$. For any distance function $d$ and any aggregation function $f$, $\mathcal{K}\models_{d,f} \psi$ if and only if $\mathcal{T}\models_{d,f} \psi$.\\
\begin{proof}
Let $\psi$ be of the form $C\sqsubseteq D$ where $C, D$ are concepts.
Let $\Sigma'=Sig(\mathcal{T}\cup \{C\sqsubseteq D\})$. We can conclude that $\mathcal{K}\models_{d,f} \psi$ if and only if $Mod^{F}_{d,f}(\mathcal{K})\subseteq Mod^{F}(\{\psi\})$ by Definition \ref{def:d-entail}. That is, for each $\Sigma$-feature $\mathcal{F}=\langle \Xi,\mathcal{H} \rangle\in Mod^{F}_{d,f}(\mathcal{K})$, $\mathcal{F} \in Mod^{F}(\{C\sqsubseteq D\})$ by Definition \ref{def:d-entail}. Then $\Xi\subseteq T_{\Sigma}(\neg C\sqcup D))$ and $\mathcal{H}$ is arbitrary since $\psi$ is a concept inclusion. Therefore, $\Lambda^{+}_{d,f}(\Pi_{\Sigma}(\mathcal{T}),T_{\Sigma}) \subseteq T_{\Sigma'}(\neg C\sqcup D)$ since for all $\Xi\in \Lambda^{+}_{d,f}(\Pi_{\Sigma}(\mathcal{T}),T_{\Sigma'})$, $\Xi\subseteq T_{\Sigma}(\neg C\sqcup D))$ by Definition \ref{def:mini-feature}. We still conclude that\\ $\Lambda^{+}_{d,f}(\Pi_{\Sigma}(\mathcal{T}),T_{\Sigma'}) \subseteq T_{\Sigma'}(\neg C\sqcup D)$ since $T_{\Sigma'}(\neg C\sqcup D)=T'_{\Sigma}(\neg C\sqcup D)$ obtained by removing all literals of $\Sigma\setminus \Sigma'$ in $T_{\Sigma}(\neg C\sqcup D)$.

Next, we claim that
\begin{center}
$\mathcal{T}\models_{d,f} C\sqsubseteq D$ if and only if $\Lambda^{+}_{d,f}(\Pi_{\Sigma}(\mathcal{T}),T_{\Sigma'}) \subseteq T_{\Sigma'}(\neg C\sqcup D)$.
\end{center}
Now, we prove this claim. If $\mathcal{K}=(\mathcal{T},\mathcal{A})$ where $\mathcal{A}=\emptyset$, then $Mod^{F}_{d,f}(\mathcal{K})=\{\mathcal{F}_{1}$, $\ldots$, $\mathcal{F}_{m}\}$ where $\mathcal{F}_{i}=\langle \Xi_{i}, \emptyset\}$ $i\in \{1,\ldots,m\}$. Let $\Xi=\bigcup^{m}_{i=1}\Xi_{i}$ where $\{\Xi_{1},\ldots,\Xi_{m}\}$ is an enumeration of all possible subsets of $\Xi_{\mathcal{K}}$. For each $\Xi_{i}$ ($1\leq i \leq m$), $\exists P \in \cup \Xi_{i}$ if and only if $\exists P^{-} \cup \Xi_{i}$ for any role name $P\in \Sigma_{R}$. Thus $\Xi = \Xi_{\mathcal{K}}$. On the other hand, analogously, $Mod^{F}(\{C\sqsubseteq D\})=\{\mathcal{F}'_{1},\ldots, \mathcal{F}'_{n}\}$ where $\mathcal{F}'_{i}=\langle \Xi'_{i},\emptyset \rangle$ ($1\leq i \leq n$) where $\{\Xi'_{1},\ldots,\Xi'_{n}\}$ is an enumeration of all possible subsets of $T_{\Sigma}(\neg C\sqcup D)$. $\Xi'=\bigcup^{n}_{i=1}\Xi'_{i}$. For each $\Xi'_{i}$ ($1\leq i \leq n$), $\exists P \in \cup \Xi'_{i}$ if and only if $\exists P^{-} \cup \Xi'_{i}$ for any role name $P\in \Sigma_{R}$. Thus $\Xi' = T_{\Sigma}(\neg C\sqcup D)$. So $Mod^{F}_{d,f}(\mathcal{K})\subseteq Mod^{F}(\{C\sqsubseteq D\})$ if and only if  $\{\mathcal{F}_{1},\ldots,\mathcal{F}_{m}\}\subseteq \{\mathcal{F}'_{1},\ldots,\mathcal{F}'_{n}\}$. That is, $\{\mathcal{F}_{1},\ldots,\mathcal{F}_{m}\}\subseteq \{\mathcal{F}'_{1},\ldots,\mathcal{F}'_{n}\}$ if and only if  $\Xi_{\mathcal{K}}\subseteq T_{\Sigma}(\neg C \sqcup D)$ since $\{\Xi_{1},\ldots,\Xi_{m}\}$ and $\{\Xi'_{1},\ldots,\Xi'_{n}\}$ are enumerations of all possible subsets of $\Xi_{\mathcal{K}}$ and $T_{\Sigma}(\neg C\sqcup D)$ respectively. Therefore, $Mod^{F}_{d,f}(\mathcal{K})\subseteq Mod^{F}(\{C\sqsubseteq D\})$ if and only if $\Xi_{\mathcal{K}}\subseteq T_{\Sigma}(\neg C \sqcup D)$. Therefore, $\mathcal{K}\models_{d,f} \psi$ if and only if $\mathcal{T}\models_{d,f} \psi$.
\end{proof}

\noindent \textbf{Proposition} \ref{prop:d-consistent}\\
Let $\Sigma$ be a signature and $\mathcal{K}=(\mathcal{T},\mathcal{A})$ a KB over $\Sigma$. For any distance function $d$ and any aggregation function $f$, let
\begin{itemize}
\item $Cn_{d,f}(\mathcal{T}) = \{ \psi$ is an inclusion $\mid \mathcal{T}\models_{d,f} \psi \}$;
\item $Cn^{\mathcal{T}}_{d,f}(\mathcal{A}) = \{ \varphi$ is an assertion $\mid (\mathcal{T},\mathcal{A})\models_{d,f} \varphi \}$.
\end{itemize}
We conclude that both $Cn_{d,f}(\mathcal{T})$ and $Cn^{\mathcal{T}}_{d,f}(\mathcal{A})$ are
consistent.\\
\begin{proof}
We only need to show that $Cn_{d,f}(\mathcal{T})\not\models \top\sqsubseteq \bot$. Assume that $Cn_{d,f}(\mathcal{T})\models \top\sqsubseteq \bot$. $Mod^{T}(Cn_{d,f}(\mathcal{T}))=Mod^{T}_{d,f}(\mathcal{T})$ since $Cn_{d,f}(\mathcal{T})$ is the deductive closure of $\models_{d,f}$ over $\mathcal{T}$. Thus $Mod^{T}_{d,f}(\mathcal{T})\subseteq Mod^{T}(\{\top\sqsubseteq \bot\})$ while $Mod^{T}(\{\top\sqsubseteq \bot\}=\emptyset$ and $Mod^{T}_{d,f}(\mathcal{T})\neq \emptyset$ by this claim in the proof of Proposition \ref{prop:Preservation-TBox}.

Suppose, for the sake of contradiction, that $Cn^{\mathcal{T}}_{d,f}(\mathcal{A})$ is inconsistent. That is, there is an assertion $C(a)$ where $C$ is a concept and $a$ is an individual name such that $Cn^{\mathcal{T}}_{d,f}(\mathcal{A})\models C(a)$ and $Cn^{\mathcal{T}}_{d,f}(\mathcal{A})\models \neg C(a)$. Then $Mod^{F}(Cn^{\mathcal{T}}_{d,f}(\mathcal{A}))=Mod^{F}_{d,f}((\mathcal{T},\mathcal{A}))$ since $Cn^{\mathcal{T}}_{d,f}(\mathcal{A})$ is the deductive closure of $\models_{d,f}$ over $\mathcal{A}$ w.r.t. $\mathcal{T}$. Thus, $(\mathcal{T},\mathcal{A})\models_{d,f} C(a)$ and $(\mathcal{T},\mathcal{A})\models_{d,f}\neg C(a)$ at the same time. Then $Mod^F_{d,f}((\mathcal{T},\mathcal{A}))\subseteq Mod^F(\{C(a)\})$ and $Mod^F_{d,f}((\mathcal{T},\mathcal{A}))\subseteq Mod^F(\{\neg
C(a)\})$. Thus $Mod^F_{d,f}((\mathcal{T},\mathcal{A}))\subseteq
Mod^F(\{C(a)\})\cap Mod^F(\{\neg C(a)\}) = Mod^F(\{C(a), \neg C(a)\}) =\emptyset$, that is, $Mod^F_{d,f}((\mathcal{T},\mathcal{A}))=\emptyset$. However, $Mod^F_{d,f}((\mathcal{T},\mathcal{A})) \neq \emptyset$ by Proposition \ref{prop:minimal-model-fearture}, we have arrived at a contradiction. 
\end{proof}

\noindent \textbf{Proposition} \ref{prop:relevance}\\
Let $\Sigma$ be a signature, $\mathcal{K}$ a KB, and, $\phi$ a non-tautology over $\Sigma$. If $d$ is an unbiased function and $Sig(\mathcal{K})\cap Sig(\{\phi\})=\emptyset$ then for any aggregation function $f$, $\mathcal{K}\not\models_{d,f} \phi$.\\
\begin{proof}
Let $\mathcal{K} = (\mathcal{T}, \mathcal{A})$. If $\phi$ is a contradiction then this claim already holds. Otherwise, let us consider three forms of $\phi$:
\begin{enumerate}
\item If $\phi$ is of the form $C\sqsubseteq D$ where $C,D$ are concepts. By Proposition \ref{prop:Preservation-TBox}, we only prove that $\mathcal{T}\not \models_{d,f} \phi$. Let $\Sigma_1 = Sig(\mathcal{T})$ and $\Sigma_2 = Sig(\phi)$. Thus $\Sigma_1 \cap \Sigma = \emptyset$ and $\Sigma_1 \cup \Sigma_2 \subseteq \Sigma$. Let $\tau_1$ be a $\Sigma_1$-type and $\tau_1 \in \Lambda^{+}_{d,f}(\Pi_{\Sigma_1}(\mathcal{T}),T_{\Sigma_1})$ and $\tau_2 \not \in T_{\Sigma_2} (\neg C \sqcup D)$. By Corollary \ref{cor:p-consistent}, $\tau_1$ exists. Because $\phi$ is neither a contradiction nor a tautology, $\tau_2$ exists. Let $\tau = \tau_1 \cup \tau_2$ (i.e., the union of $\tau_1$ and $\tau_2$) since $\tau_1 \cap \tau_2 = \emptyset$. 
For each $i$ ($1 \le i \le m$), for any basic concept $B$ occurring in $\neg C_i \sqcup D_i$, $B \in \tau_1$ if and only if $B \in \tau$. Because $d$ is unbiased, we can conclude that $d(\tau_1,T_{\Sigma}(\neg C_{i}\sqcup D_{i}))=d(\tau,T_{\Sigma}(\neg C_{i}\sqcup D_{i}))$ for any $i\in \{1,\ldots,m\}$. Because $\tau_1 \in \Lambda^{+}_{d,f}(\Pi_{\Sigma}(\mathcal{T}),T_{\Sigma})$, we can conclude that $\tau \in \Lambda^{+}_{d,f}(\Pi_{\Sigma}(\mathcal{T}),T_{\Sigma})$. However, $\tau\not\in T_{\Sigma}(\neg C\sqcup D)$ since $\tau_2 \not \in \in T_{\Sigma}(\neg C\sqcup D)$ and $\tau_2 \subseteq \tau$. Therefore, $\Lambda^{+}_{d,f}(\Pi_{\Sigma}(\mathcal{T}),T_{\Sigma}) \not \subseteq T_{\Sigma}(\neg C\sqcup D)$, that is, $\mathcal{T} \not\models_{d,f} \phi$. By Proposition \ref{prop:Preservation-TBox}, we can conclude that $\mathcal{K} \not\models_{d,f} \phi$.

\item If $\phi$ is of the form $C(a)$ where $C$ is a concept and $a$ is an individual name. Let $\Sigma_1 = Sig(\mathcal{T})$ and $\Sigma_2 = Sig(\phi)$. Thus $\Sigma_1 \cap \Sigma = \emptyset$ and $\Sigma_1 \cup \Sigma_2 \subseteq \Sigma$. Let $\tau_1$ be a $\Sigma_1$-type and $\tau_1 \in \Lambda^{+}_{d,f}(\Pi_{\Sigma_1}(\mathcal{T}),T_{\Sigma_1})$ and $\tau_2 \not \in T_{\Sigma_2} (C)$. Because $C(a)$ is neither a contradiction nor a tautology, $\tau_2$ exists. Let $\mu =\mu_1 \cup \mu_2$. By the proof of (1), if $d$ is unbiased then for any aggregation $f$, we can conclude that $\mu \in \Lambda^{+}_{d,f}(\Pi_{\Sigma}(\mathcal{T}),T_{\Sigma})$. However, $\tau$ does not satisfy $C$. Then those features of the form $(\Xi, \mathcal{H})$ with $\tau \in \Xi$ do not satisfy $C$. Such a feature always exists since $\tau_1$ is a arbitrary type. Therefore, $\mathcal{K} \not \models_{d,f} \phi$. 
\item If $\phi$ is of the form $P(a,b)$ where $P$ is a role name and $a,b$ are individual names. Let $\mathcal{F} = (\Xi, \mathcal{H})$ be a feature in $Mod^F_{d,f}(\mathcal{K})$. Let $\mathcal{F}'$ be a new feature obtained from $\mathcal{F}$ by removing $P(a, b)$ or $P^-(b, a)$ in $\mathcal{H}$. Since $Sig(\mathcal{K}) \cap Sig(P(a, b)) = \emptyset$, we can still conclude that $\mathcal{F}' \in Mod^F_{d,f}(\mathcal{K})$ while $\mathcal{F}'$ does not satisfy $P(a, b)$. Therefore, $\mathcal{K} \not \models_{d,f} \phi$. We can analogously prove the form $P^-(a, b)$. 
\end{enumerate}
Based on (1), (2) and (3), we conclude that $\mathcal{K}\not\models_{d,f} \phi$.
\end{proof}

\noindent \textbf{Proposition} \ref{prop:nonmonotonic}\\
For any distance function $d$ and any aggregation function $f$, $\models_{d,f}$ is non-monotonic.\\
\begin{proof}
Let $\Sigma = \{A, a\}$. We can conclude that $\{A(a)\}\models_{d,f} A(a)$ and $\{\neg A(a)\}\models_{d,f} \neg A(a)$ by Proposition \ref{prop:d-consistent}. However, $\{A(a),\neg A(a)\}\not\models_{d,f} A(a)$ and $\{A(a),\neg A(a)\}\not\models_{d,f} \neg A(a)$ by Definition \ref{def:d-entail}.
\end{proof}

\noindent \textbf{Proposition} \ref{prop:partition}\\
Let $\Sigma$ be a signature and $\mathcal{K}$ a KB over $\Sigma$. Assume that $\mathcal{K}=\mathcal{K}'\oplus \mathcal{K}''$ where $\mathcal{K}'$ is consistent. For each axiom $\phi$ with $Sig(\phi)\cap Sig(\mathcal{K}'')=\emptyset$, if $\mathcal{K}'\models \phi$ then for any distance function $d$ and any hereditary aggregation function $f$, $\mathcal{K}\models_{d,f}\phi$.\\
\begin{proof}
If $\phi$ is a tautology then we can directly conclude that $\mathcal{K}\models_{d,f}\phi$. Otherwise, $\phi$ is a non-tautology. For each feature $\mathcal{F} = (\Xi, \mathcal{H}) \in Mod^F_{d,f}(\mathcal{K})$, $\Xi = \Xi' \cup \Xi''$ and $\mathcal{H} = \mathcal{H}' \cup \mathcal{H''}$ for some feature $\mathcal{F}' = (\Xi', \mathcal{H}') \in Mod^F_{d,f}(\mathcal{K}')$ and some feature $\mathcal{F}'' = (\Xi'', \mathcal{H}'') \in Mod^F_{d,f}(\mathcal{K}'')$ by Definition \ref{def:mini-feature} since $\mathcal{K}=\mathcal{K}'\oplus \mathcal{K}''$. Because $\mathcal{K}'$ is consistent, we have $\mathcal{K}'\models \phi$ if and only if $\mathcal{K}'\models_{d,f} \phi$ by Proposition \ref{prop:Consistent-P}. That is, $\mathcal{F}' \in Mod^{F}_{d,f}(\mathcal{K}')=Mod^{F}(\mathcal{K}') \subseteq Mod^F(\{\phi\})$. Because $Sig(\phi)\cap Sig(\mathcal{K}'')=\emptyset$, $\mathcal{F} \in Mod^F(\{\phi\})$. Therefore, $\mathcal{K}\models_{d,f} \phi$.
\end{proof}

\noindent \textbf{Proposition} \ref{prop:cautious}\\
For any distance function $d$ and any monotonic hereditary aggregation function $f$, $\models_{d,f}$ is cautious.\\
\begin{proof}
Let us prove that $\models_{d,f}$ satisfies three properties: cautious reflexivity, cautious monotonicity, and, cautious cut.
\begin{itemize}
\item
If $\mathcal{K}=\mathcal{K}'\oplus \mathcal{K}''$ and $\mathcal{K}'$ is consistent, then, for any distance function $d$ and any monotonic hereditary aggregation function $f$, $\mathcal{K}\models_{d,f}\varphi$ since $\mathcal{K}\models_{d,f} \varphi$ for all axiom $\varphi\in \mathcal{K}'$ by Proposition \ref{prop:partition}. 
\item
If $\mathcal{K}\models_{d,f} \varphi$ then $Mod^F_{d,f}(\mathcal{K}) \subseteq Mod^F(\{\varphi\}) \subseteq Mod^F_{d,f}(\{\varphi\})$ by Definition \ref{def:d-entail} and Proposition \ref{prop:minimal-model-fearture}. If $\mathcal{K}\models_{d,f} \psi$ then $Mod^F_{d,f}(\mathcal{K})\subseteq \mathcal{F} \in Mod^F(\{\psi\})$ by Definition \ref{def:d-entail}. $Mod^F_{d,f}(\mathcal{K} \cup \{\varphi\}) \subseteq Mod^F_{d,f}(\mathcal{K})$ since $Mod^F_{d,f}(\mathcal{K}) \cap Mod^F_{d,f}(\{\varphi\}) \neq \emptyset$ by Definition \ref{def:mini-feature}. Then $Mod^F_{d,f}(\mathcal{K} \cup \{\varphi\}) \subseteq Mod^F(\{\psi\})$, that is, $\mathcal{K} \cup \{\varphi\}\models_{d,f} \psi$. 
\item
If $\mathcal{K}\models_{d,f} \varphi$ then $Mod^F_{d,f}(\mathcal{K}) \subseteq Mod^F(\{\varphi\}) \subseteq Mod^F_{d,f}(\{\varphi\})$ by Definition \ref{def:d-entail} and Proposition \ref{prop:minimal-model-fearture}. If $\mathcal{K} \cup \{\varphi\}\models_{d,f} \psi$ then $Mod^F_{d,f}(\mathcal{K} \cup \{\varphi\}) \subseteq Mod^F(\{\psi\})$ by Definition \ref{def:d-entail}. $Mod^F_{d,f}(\mathcal{K} \cup \{\varphi\}) \supseteq  Mod^F_{d,f}(\mathcal{K}) \cap Mod^F_{d,f}(\{\varphi\}) = Mod^F_{d,f}(\mathcal{K})$ since $Mod^F_{d,f}(\mathcal{K}) \cap Mod^F_{d,f}(\{\varphi\}) \neq \emptyset$ by Definition \ref{def:mini-feature}. Then, $Mod^F_{d,f}(\mathcal{K}) \subseteq Mod^F(\{\psi\})$, that is, $\mathcal{K} \models_{d,f} \psi$.
\end{itemize}
Based on (1), (2) and (3), we conclude that $\models_{d,f}$ is cautious by the definition of cautious relation.
\end{proof}

\end{document}